\documentclass[acmtog]{acmart}

\usepackage{amsthm}
\usepackage{color,hyphenat,balance}
\usepackage[fleqn,tbtags]{mathtools}
\usepackage{autonum}
\usepackage{overpic}
\usepackage{wrapfig}
\usepackage{contour}
\usepackage{microtype}
\usepackage{multirow}
\usepackage{soul}
\usepackage{algorithm}
\usepackage{xspace}
\usepackage{algpseudocode}
\usepackage{diagbox}

\usepackage{mathtools}
\usepackage{lineno}

\acmSubmissionID{472} 

\copyrightyear{2024}
\acmYear{2024}
\setcopyright{acmlicensed}\acmConference[SA Conference Papers '24]{SIGGRAPH Asia 2024 Conference Papers}{December 3--6, 2024}{Tokyo, Japan}
\acmBooktitle{SIGGRAPH Asia 2024 Conference Papers (SA Conference Papers '24), December 3--6, 2024, Tokyo, Japan}
\acmDOI{10.1145/3680528.3687610}
\acmISBN{979-8-4007-1131-2/24/12}

\begin{document}

\title{Neural Garment Dynamic Super-Resolution}


\author{Meng Zhang}
\authornote{M. Zhang is the corresponding author.}
\email{lynnzephyr@gmail.com}
\orcid{0000-0003-2384-0697}
\affiliation{%
 \institution{School of Computer Science and Engineering, Nanjing University of Science and Technology}\country{China}}

\author{Jun Li}
\email{junli@njust.edu.cn}
\orcid{0000-0003-3716-671X}
\affiliation{%
 \institution{School of Computer Science and Engineering, Nanjing University of Science and Technology}\country{China}}






\renewcommand\shortauthors{Zhang M. et al}


\begin{abstract}
Achieving efficient, high-fidelity, high-resolution garment simulation is challenging due to its computational demands. Conversely, low-resolution garment simulation is more accessible and ideal for low-budget devices like smartphones. In this paper, we introduce a lightweight, learning-based method for garment dynamic super-resolution, designed to efficiently enhance high-resolution, high-frequency details in low-resolution garment simulations.
Starting with low-resolution garment simulation and underlying body motion, we utilize a mesh-graph-net to compute super-resolution features based on coarse garment dynamics and garment-body interactions. These features are then used by a hyper-net to construct an implicit function of detailed wrinkle residuals for each coarse mesh triangle. Considering the influence of coarse garment shapes on detailed wrinkle performance, we correct the coarse garment shape and predict detailed wrinkle residuals using these implicit functions. Finally, we generate detailed high-resolution garment geometry by applying the detailed wrinkle residuals to the corrected coarse garment.
Our method enables roll-out prediction by iteratively using its predictions as input for subsequent frames, producing fine-grained wrinkle details to enhance the low-resolution simulation. Despite training on a small dataset, our network robustly generalizes to different body shapes, motions, and garment types not present in the training data. We demonstrate significant improvements over state-of-the-art alternatives, particularly in enhancing the quality of high-frequency, fine-grained wrinkle details.
\textit{Code and data is released in \href{https://github.com/MengZephyr/Neural-Garment-Dynamic-Super-resolution/}{https://github.com/MengZephyr/Neural-Garment-Dynamic-Super-resolution/}}

\end{abstract}

\ccsdesc[500]{Computing methodologies~Physical simulation}
\ccsdesc[500]{Computing methodologies~Mesh geometry models}
\ccsdesc[500]{Networks}
       


\keywords{garment dynamics, super resolution, high frequency detail, animation, generalization}


\begin{teaserfigure}
\centering
\includegraphics[width=0.95\textwidth]
{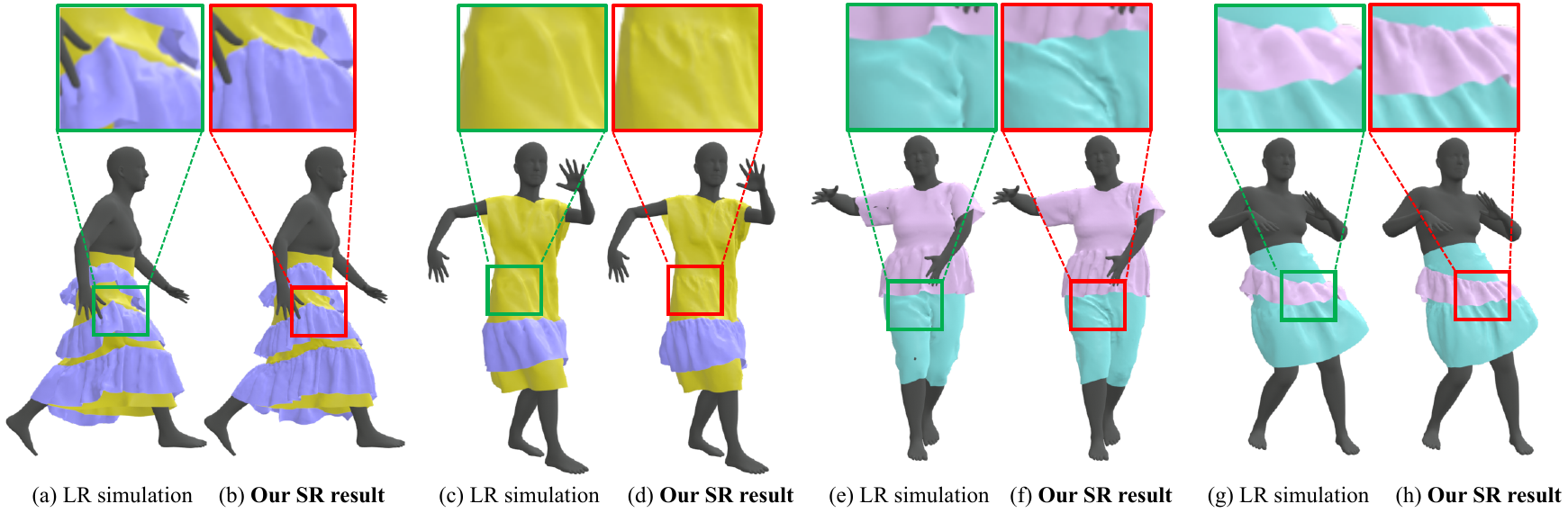}
\vskip -0.15in
\caption{
Given low-resolution (LR) garment simulation as input (a,c,e,g), we present a learning-based method of \emph{garment dynamic super resolution} to construct high-resolution garment dynamic geometry (b,d,f,h). 
We learn super resolution (SR) features from the coarse garment dynamics and the garment-body interactions, to produce high-resolution fine-grained and plausible wrinkle details to enhance the low-resolution geometry of various garment types, and enable the generalization ability of our method across unseen body motions (b,d,f,h), unseen body shape (d,f), and unseen garment types (f,h).
}  
\label{fig:teaser}
\end{teaserfigure}


\maketitle

\section{Introduction}

Realistic garment behavior is crucial across various applications, including virtual try-on, 3D character design for games, movies, and VR/AR.
To reproduce real garments in virtual world, physics-based simulation of garment dynamics has been extensively studied in the graphic community for decades, leading to substantial and promising progress \cite{10.1145/3450626.3459767, 10.1145/3610548.3618157}. One key challenge here is achieving high-resolution results and reproducing the intricate wrinkle details caused by complex human-garment interactions, which are essential for high visual quality. Since garments are typically represented as meshes in simulations, increasing resolution helps to model garment geometry with fine details~\cite{10.1145/3450626.3459787}. However, this approach results in computationally expensive simulations and difficulties in efficient data transmission and storage for data sequences.
Although advancements in graphics hardware have significantly improved the run-time efficiency and fidelity of physics-based simulations, these improvements are not designed for low-budget application scenarios.
Recently, deep-learning-based algorithms have emerged as efficient alternatives for producing garment dynamics, owing to their relatively lower computational complexity. A popular approach involves decomposing garment deformation into high- and low-frequency components \cite{patel2020tailornet, santesteban2019learning, 10.1145/3528233.3530709, bertiche2022neural}. Several studies further enhance this method by using normal maps as intermediate representations to augment up-sampled low-frequency garment dynamic meshes \cite{lahner2018deepwrinkles, zhang2021deep}. Halimi et al.~\shortcite{halimi2023physgraph} and Chen et al.~\shortcite{chen2023deep} leverage mesh-based graph neural networks to refine the high-resolution meshes, up-sampled from coarse garments, by adding fine wrinkle details. However, these methods still encounter challenges related to the time consumption and geometric complexity of processing deformations on high-resolution garment meshes. 

In this work, we formulate efficiently producing high-fidelity high-resolution garment geometry as a super-resolution task, aimed at either compressing high-resolution garment data or speeding up simulations for fine wrinkle details. 
This involves reconstructing high-frequency wrinkle details based on the coarse garment geometry, as low-resolution simulation is easily accessible and suitable for the application requirement of low-budget scenarios.  
To this end, we propose neural garment dynamic super-resolution (\emph{GDSR}). Using low-resolution garment simulations as input, our method processes dynamic garment geometry from low to high resolution in an end-to-end manner, bypassing the need for intermediate normal maps used in other approaches (e.g., \cite{zhang2021deep, lahner2018deepwrinkles}), which introduce additional computation for normal-guided deformation to generate high-resolution geometry. 

Specifically, \emph{GDSR} uses a mesh-graph-net to extract super resolution features from the coarse garment dynamics, as well as the garment interactions both with the body and among fabric layers. It further exploits the super-resolution features, using a hyper-net to construct continuous up-sampling fields to generate fine detailed high-resolution garment geometry. Considering the impact of coarse garment shape on the detail wrinkle performance, \emph{GDSR} effectively corrects the low-resolution garment geometry, by decoding the super-resolution features, while simultaneously predicting high-resolution wrinkle details. 
To ensure temporal coherence in wrinkle details, \emph{GDSR} performs the iterative roll-out prediction with the resultant corrected coarse garment as a historical reference. 


Our method generalizes across various motion dynamics, body shapes, and garment types, even those not present in the training data.
We demonstrate this generalization ability in Section \ref{sec: results}. Comparing our approach to \cite{zhang2021deep,halimi2023physgraph}, we show that our method enhances garment dynamics with more vivid geometric wrinkle details.
The \emph{GDSR} network is lightweight, with a storage size of only 65MB, independent of the garment mesh size. It achieves high running efficiency, potentially enabling fine-detailed high-resolution garment simulation in real-time.
Computation times are discussed in Section \ref{subsec: runningtime} and the supplementary material.

Our key contributions are:
\begin{itemize}
\item \emph{dynamic super-resolution features} to encode low-resolution garment dynamics and garment-body interactions.
\item \emph{wrinkle detail inference} to synthesize high-resolution wrinkle details while correcting low-resolution garment shape.
\item \emph{generalization} across various body motions, body shapes, and garment types.
\end{itemize}

\section{Related Work}
\paragraph{Physics-based optimization}
Physics-based garment simulation has been a fascinating topic \cite{choi2005research, nealen2006physically, kim2020dynamic, 10.1007/978-3-642-21799-9_35, 10.1145/566654.566624, liu2013fast, 10.1145/280814.280821, cirio2014yarn} in the graphics community for decades.
But computation cost and stability concern is still the major issue, 
especially in the high-resolution garment simulation.
Alternative approaches have been explored, to enhance fine wrinkle details by optimizing the coarse garment geometry based on physics constraint \cite{10.2312:SCA:SCA10:085-091, 10.1145/1198555.1198573}, operating on the garment mesh with stretch tensors \cite{rohmer2010animation, gillette2015real, li2018foldsketch, chen2021fine}, quadratic shape functions \cite{remillard2013embedded}, grid structure \cite{10.1145/3450626.3459787}, {and customized hierarchical prolongation operators \cite{zhang2023progressive}}. Notably, Wang et al.\shortcite{10.1145/3450626.3459787} achieved detail synthesis at sub-millimeter levels based on a computing environment equipped with powerful GPUs. 
{However, these approaches still remain computationally intensive.}

\paragraph{Data-driven synthesis}
Data-driven techniques are naturally explored to synthesize thin shell deformations and wrinkle details \cite{frohlich2011example, grinspun2003discrete}.
Example-based method is applied to synthesize wrinkle details by blending the nearest wrinkle samples, searching based on body features that affect the garment \cite{Wang2010, xu2014sensitivity}, or garment geometry local features \cite{Feng2010, zurdo2012animating}.
DRAPE ~\cite{guan2012drape} learns linear subspace models of garment deformations with respect to the body pose variation. 
The following work ~\cite{hahn2014subspace} uses adaptive bases to achieve richer wrinkle details. 
Kavan et al.~\shortcite{Kavan2011} learned linear up-sampling operators, however, which is limited to the case of simple coarse cloth.
Our work is loosely related to the idea of learning up-sampling operator. In contrast to previous works, we achieve efficient wrinkle detail enhancement on complex stylized garment types.

\paragraph{Neural-based prediction}
Recent research focuses on using neural networks for predicting garment deformations, bridging garment dynamics and body motions \cite{d2022n, gundogdu2019garnet, li2023ctsn, wang2019learning, 10.1145/3528233.3530709, bertiche2022neural}, where many studies rely on the SMPL model \cite{10.1145/2816795.2818013}.
A common approach decomposes garment deformation into low- and high-frequency components, with low-frequency driven by static body poses and high-frequency as displacement predictions. Mesh-based graph networks \cite{pfaff2020learning} and their successors \cite{libao2023meshgraphnetrp, grigorev2022hood} use physics-based constraints for improved accuracy. UV parameterization in 2D space is also explored \cite{10.1145/3478513.3480497, 10.1145/3550454.3555485, shen2020gan, shao2023towards}.
Despite advancements, these methods struggle with high-resolution garment deformations due to time consumption and geometric complexity. Recent efforts include untangling multi-layered garments via untangled fields \cite{santesteban2022ulnef} and GNN-based models for fabric layer interactions \cite{lee2023clothcombo}, though they remain limited to static body poses.
Since expensive the high-resolution garment simulation is, several research works mainly focus on generating high-resolution wrinkle details to enhance the coarse garment geometry. Later et al.~\shortcite{lahner2018deepwrinkles} employed a conditional generative adversarial network to add high-frequency details to the normal maps of the garments. Zhang et al.~\shortcite{zhang2021deep} cast the detail enhancement task as a style transfer of the normal maps. Both of those works adopt the normal map to represent the garment geometry at a cost of time, that spends on normal map rendering and normal guided geometry deformation.
The architecture of mesh-based graph net is applied in \cite{halimi2023physgraph} to optimize the high-resolution wrinkle detail geometry, by iterative running the graph network to solve the physically-based potential energy. 
Chen et al.~\shortcite{chen2023deep} trained garment-specific neural networks to translate the detailed garment dynamics from the coarse garment sequence. To our knowledge, none of those previous works produce garments with high-frequency details as the pleated lace and skirt shown in Figure \ref{fig:teaser}. 

\paragraph{Super resolution.}
Techniques of deep learning have made significant progress for the super-resolution task of image and video \cite{ledig2017photo, zhang2018residual, haris2019recurrent, sajjadi2018frame, chu2020learning, Chen2021implicitSR}. 2D convolution neural networks (CNN) are commonly used, which are naturally compatible with the data structure of images, and help achieve a hierarchical feature encoding and decoding.
Chen et al. \shortcite{chen2021multi} adopted the CNN-based super-resolution technique to synthesize cloth wrinkles by recording multiple garment geometry features in 2D image space, where the detail representation capability is limited by the image size. In contrast, we feature our garment dynamic super-resolution in the original low-resolution garment mesh with no conversion between the geometry and image space, and construct continuous up-sampling fields locally for coarse mesh triangles to produce high-resolution detailed garment geometry.
Additionally, different from the image/video-based super-resolution operating on static low-resolution content, our work further explores to correct the coarse content (garment), jointly with the high-frequency detail reconstruction.


\section{Overview}
Starting from the low-resolution garment dynamics $\{C_t\}$ and the underlying body motions $\{B_t\}$, our \emph{neural garment dynamic super-resolution} (\emph{GDSR}) adeptly generates a high-resolution garment geometry sequence ${G_t}$ imbued with vibrant fine wrinkle details. 
\begin{figure}[ht]
\includegraphics[width=0.97\columnwidth]{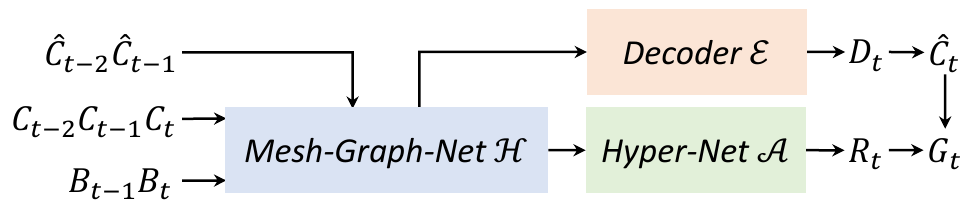}
   \vskip -0.1in
   \centering
   \caption{
    \textbf{\emph{GDSR} overview.} With the low-resolution garment dynamics $C_{t-2}$, $C_{t-1}$ and $C_t$, the underlying body motion $B_{t-1}$, $B_t$, and the previous prediction $\hat{C}_{t-2}$, $\hat{C}_{t-1}$, our \emph{GDSR} predicts the displacements $D_t$ to compute the corrected coarse garment shape $\hat{C}_t$, and jointly generates detail wrinkle residual $R$ to produce the final high-resolution garment geometry $G_t$.
  }
\label{fig:GDSR-B}
\vskip -0.05in
\end{figure}

\begin{figure*}[t!]
    \centering
    \includegraphics[width=0.95\textwidth]{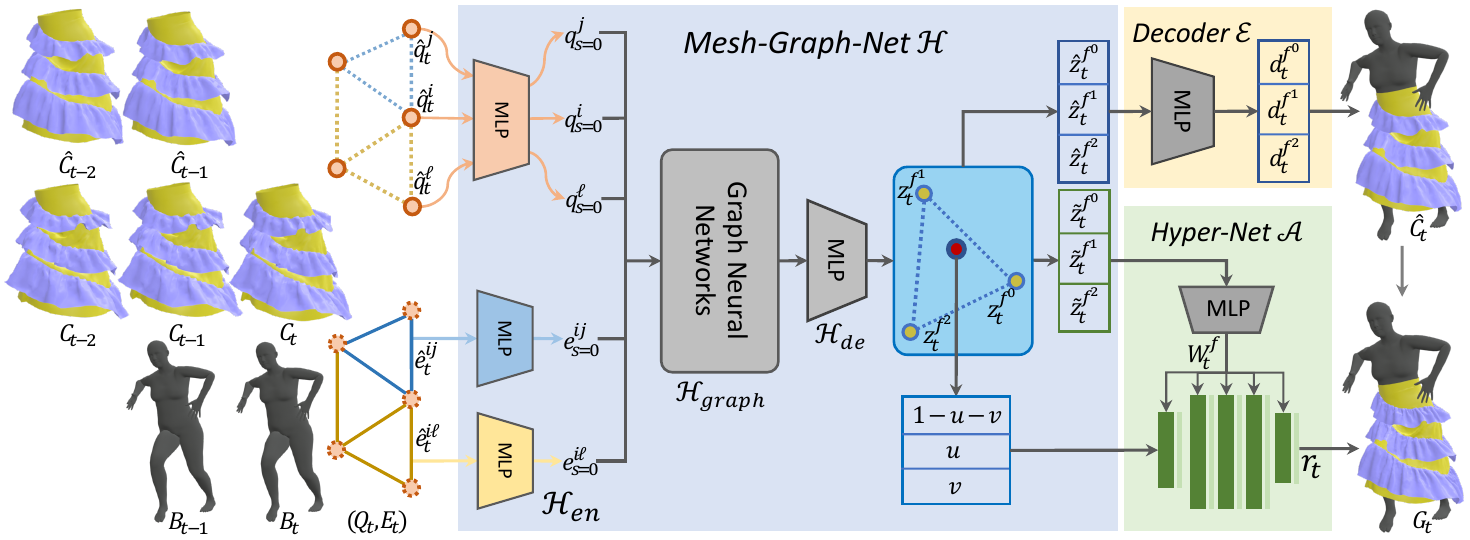}
    \vskip -0.1in
    \caption{\textbf{Garment dynamic super-resolution architecture.} 
    Our method operates in a roll-out prediction manner. 
    We represent the inputs of low-resolution simulations, body motions, and downsampled previous predictions as a graph with node features $Q_t$ and edge features $E_t$.
    First, we employ \emph{Mesh-Graph-Net} $\mathcal{H}$ to compute super-resolution features $z_t:=[\hat{z}_t | \tilde{z}_t]$ for coarse garment mesh. 
    Considering the impact of coarse garment shape on detail wrinkle performance, \emph{Decoder} $\mathcal{E}$ decodes features $\hat{z}_t$ to correct the low-resolution garment shape, resulting in $\hat{C}_t$. 
    Simultaneously, \emph{Hyper-Net} $\mathcal{A}$ exploits super-resolution features $\tilde{z}_t$ to construct an implicit function $W_t^f$ of detail wrinkle residuals for each mesh triangle. Finally, we generate the high-resolution garment $G_t$ by applying the detail wrinkle residuals on the corrected coarse garment. 
    }
    \label{fig:networks}
    \vskip -0.1in
\end{figure*}

We introduce \emph{GDSR} to predict the displacement $D_t$ to correct the coarse garment shape of the low-resolution simulation, and meanwhile, synthesize wrinkle detail residuals $R_t$. As shown in Figure \ref{fig:GDSR-B}, the network of \emph{GDSR} comprises three integral network blocks: {Mesh-Graph-Net} $\mathcal{H}$ (refer to Section \ref{subsec: net H}), {Decoder} $\mathcal{E}$ (refere to Section \ref{subsec: net C}), and {Mesh-Hyper-Net} $\mathcal{A}$ (refer to Section \ref{subsec: net A}). 
The super-resolution features for all low-resolution garment vertices are computed using $\mathcal{H}$. Recognizing the significant impact of coarse garment shape on wrinkle detail performance, we rectify the coarse garment geometry at the low-resolution scale by using $\mathcal{E}$ to decode the displacement $D_t$. We add the displacement $d_t^i$ to each coarse garment vertex $c_t^i$ as $\hat{c}_t^i=c_t^i + d_t^i$, resulting in the corrected coarse garment shape $\hat{C}_t$.
Simultaneously, {Mesh-Hyper-Net} $\mathcal{A}$ constructs continuous up-sampling fields for triangles of the coarse garment mesh, where wrinkle detail residuals $R_t$ are decoded in terms of the local coordinate of each triangle. Finally, the high-resolution, finely detailed garment $G_t$ is computed by combining the up-sampling of the $\hat{C}_t$ with the wrinkle detail residuals $R_t$. To account for the historical dynamics influencing the current state of the garment, we run our method in a roll-out prediction manner, incorporating $\hat{C}_{t-1}$ and $\hat{C}_{t-2}$ down-sampled from previous predictions. The detailed architecture, illustrated in Figure \ref{fig:networks}, is described in the next section.

\section{Algorithm}

Given the high-resolution $G$ and the low-resolution $C$ of a garment, we observe a commonality in their semantic 2D patterns within the UV space. Leveraging the alignment in the 2D UV space, we can establish a correspondence between the low- and high-resolution garment meshes.
Each interior point within a mesh triangle $f$ can be expressed as a sampling $\mathcal{S}$ through a linear combination of the barycentric coordinates $(u,v)$ and the three triangle vertices $c^{f^0}$, $c^{f^1}$, $c^{f^2}$. Consequently, the up-sampling of a high-resolution garment mesh from the low-resolution $C$ involves representing every vertex of the up-sampled mesh as:
\begin{equation}
    \mathcal{S}[C, f,u, v]=(1-u-v)c^{f^0}+uc^{f^1}+vc^{f^2},
\end{equation}
where $f$ is derived from the low-resolution mesh $C$. This geometric parametrization is established in the canonical space once a garment is set up and remains universally applicable across various sequences of garment dynamics. 

\subsection{Graph Feature Initialization.}
Given the low-resolution garment dynamics $C_t$, $C_{t-1}$, $C_{t-2}$, and the underlying body motion $B_t$, $B_{t-1}$, we conceptualize a graph $(Q_t, E_t)$ at current time $t$, where we assign the coarse garment node features ${\hat{q}^i}$ and edge features ${\hat{e}^{ij}}$ to represent garments based on the geometric mesh topology, where $i$ and $j$ are garment nodes connected by an edge. For multi-layered garments, we introduce additional edge features $\hat{e}^{i\ell}$ to describe the interaction between two garment nodes,  $i$ and $\ell$, each node located in a different cloth layer.  

\paragraph{Node feature initialization.}
Considering the information of garment's geometry and dynamics, we initialize the graph node feature $\hat{q}_t^i$ with the vertex normal $n_t^i$, velocity $\dot{c}_t^i$ and acceleration $\ddot{c}_t^i $ of the coarse garment vertex $c_t^i$. Additionally, in contrast to the normal-based methods \cite{lahner2018deepwrinkles, zhang2021deep} of detail enhancement, we observe that the garment-and-body interactions significantly influence the generation of garment folds and fine-grained wrinkles. Thus, we detect the garment-and-body interaction by searching the nearest body vertex $b_t$ for each coarse garment vertex $c_t^i$ using the signed distance filed estimated through ray casting \cite{roth1982ray}. We define the interaction vector $p_t^i$ as:
\begin{equation}
    p_t^i:= \mathbf{ReLU}[\sigma_b-(c_t^i-b_t)\cdot n_t^b]n_t^b,
\end{equation}
where $n_t^b$ is the normal of body vertex $b_t$, $\mathbf{ReLU}$ is the rectified linear unit function \cite{nair2010rectified}, and $\sigma_b$ is a predefined constant value. We also consider the interaction between the coarse garment vertex $c^i_{t-1}$ and the body vertex $b^i_t$ which potentially drives the garment deformation at time $t-1$, and define the interaction vector $\hat{p}_t^i$ as:
\begin{equation}
    \hat{p}_t^i:= \mathbf{ReLU}[\sigma_b-(c_{t-1}^i-b_t)\cdot n_t^b]n_t^b.
\end{equation}
Considering the history state of the garment dynamics influencing on the current state of the garment, we propagate the displacement $d^i_{t-1}=\hat{c}^i_{t-1}-c^i_{t-1}$ and the velocity $\dot{\hat{c}}^i_{t-1}$ involving the coarse garment geometries $\hat{C}_{t-1}$ and $\hat{C}_{t-2}$, down-sampled from the history high-resolution garment predictions, to perform the iterative prediction at current time $t$.
Consequently, we initialize the coarse garment node feature $\hat{q}_t^i$ as:
\begin{equation}
    \hat{q}_t^i:=[n_t^i, \dot{c}_t^i, \ddot{c}_t^i, p_t^i, \|p_t^i\|, \hat{p}_t^i, \|\hat{p}_t^i\|, d_{t-1}^i, \dot{\hat{c}}^i_{t-1}],
\end{equation}
where $\|p_t^i\|$ and $\|\hat{p}_t^i\|$ is the length of $p_t^i$ and $\hat{p}_t^i$.

\paragraph{Edge feature initialization.}
The garment edge feature $\hat{e}_t^{ij}$ between node $i$ and node $j$, is defined by the edge vector $c^j-c^i$ and edge length $\|c^j-c^i\|$ at the current and previous time $t, t-1$, and the edge length $\|c_r^i-c_r^j\|$ at the rest state $o$:
\begin{equation}
    \hat{e}_t^{ij}:=[c_t^j-c_t^i,\|c_t^j-c_t^i\|,c_{t-1}^j-c_{t-1}^i,\|c_{t-1}^j-c_{t-1}^i\|, \|c_o^i-c_o^j\|].
\end{equation}
For garment nodes $i$ and $\ell$ located in different cloth layers, we incorporate an extra edge feature $\hat{e}_t^{i\ell}$ when $\|c_t^i-c_t^{\ell}\| < \sigma_\ell$. 
This feature is defined similarly to the mesh topology-based edge feature $\hat{e}_t^{ij}$, but replaces the rest state edge length $\|c_o^i-c_o^j\|$ with the predefined constant value $\sigma_\ell$.


\subsection{Garment Super Resolution Feature Using $\mathcal{H}$} \label{subsec: net H}
%
With the initialized graph as input, we train Mesh-Graph-Net $\mathcal{H}$  to encode super-resolution features on the low-resolution garment mesh through three successive phases: Feature Encoder $\mathcal{H}_{en}$, Graph Process $\mathcal{H}_{graph}$ and Feature Decoder $\mathcal{H}_{de}$.



\paragraph{Feature Encoder $\mathcal{H}_{en}$.}
With the initialized graph $(Q,E)$ (temporal notation $t$ is omitted for a brevity of description), we employ three blocks of multi-layer perceptrons (MLP) within the Feature Encoder $\mathcal{H}_{en}$ to project the features of garment nodes $\{\hat{q}^i\}$, garment edges $\{\hat{e}^{ij}\}$, and cloth layered edges ${\hat{e}^{i\ell}}$ into three respective latent spaces, denoted as $\{q_0^i\}$, $\{e_0^{ij}\}$, and $\{e_0^{i\ell}\}$, each having the same dimension. These encoded features serve as the initialized state before undergoing the subsequent graph processing network. In our experiments, we have chosen a graph latent space dimension of $128$.

\paragraph{Graph Process $\mathcal{H}_{graph}$.} We adopt a fixed set of steps, $S$, for message passing in the Graph Process network $\mathcal{H}_{graph}$. In each block, we leverage MLPs with resiudal connection to update the graph features. Following \cite{pfaff2020learning}, we first update the edge features by passing the node features to their connected edges:
\begin{equation}
    e_s^{ij}=\mathcal{F}_{q \rightarrow e} [e^{ij}_{s-1}, q^i_{s-1}, q^j_{s-1}]+e_{s-1}^{ij},
\end{equation}
where $\mathcal{F}_{q \rightarrow e}$ represents an MLP. Second, we proceed to update the node features by passing the edges features to their corresponding adjacent nodes:
\begin{equation}
    q_s^i=\mathcal{F}_{e \rightarrow q} [q^i_{s-1}, \sum_j{e^{ij}_s}, \sum_\ell{e^{i\ell}_s}]+q^i_{s-1},
\end{equation}
where $\mathcal{F}_{e \rightarrow q}$ denotes another MLP. If there is no layered edge connection with garment node $i$, we regard $\sum_\ell{e^{i\ell}_s}$ to be $0$. 
After completing $S$ steps of message passing within $\mathcal{H}_{graph}$, we obtain the output node features denoted as $\{q_{S}^i\}$.


To expedite the propagation of features across the garment mesh, we further extend Graph Process Network $\mathcal{H}_{graph}$ to incorporate hierarchical message passing, as proposed in \cite{grigorev2022hood}. In our experiments, we set the number of message passing steps $S$ to be $6$, and establish a $2$-ring hierarchical graph construction. For more details of the hierarchical graph process, please refer to \cite{grigorev2022hood}. 


\paragraph{Feature Decoder $\mathcal{H}_{de}$.} Utilizing the node features $\{q_{S}^i\}$ as input, we apply Feature Decoder $\mathcal{H}_{de}$ to decode the ultimate super-resolution features $\{z^i\}$ for every vertex on the low-resolution garment mesh. This is achieved by employing an MLP to increase the dimensionality of the node features from $128$ to $640$. Subsequently, we segment the super resolution feature $z^i = [\hat{z}^i | \tilde{z}^i]$ into displacement super-resolution feature $\hat{z}^i$ and detail super-resolution feature $\tilde{z}^i$ with the dimensions of $128$ and $512$, respectively.   

\subsection{Coarse Garment Correction Using $\mathcal{E}$} \label{subsec: net C}
To rectify the coarse garment shape, we employ a Decoder $\mathcal{E}$ to compute the displacements $D$ with respect to the vertices of low-resolution mesh. 
Using three displacement super-resolution features $\hat{z}^{f^0}$, $\hat{z}^{f^1}$, $\hat{z}^{f^2}$ from a low-resolution mesh triangle, we decode the displacements of the three triangle vertices under the local coordinate $H^f$:
\begin{equation}
    [\tilde{d}^{f^0}, \tilde{d}^{f^1}, \tilde{d}^{f^2}] = \mathcal{E}[\hat{z}^{f^0}, \hat{z}^{f^1}, \hat{z}^{f^2}],
\end{equation}
where $\mathcal{E}$ is an MLP. Then we compute
the displacement vector $d$ of a low-resolution garment vertex $c$, by averaging the displacements of the corresponding vertices $f^c$ in the adjacent triangles $f$:
\begin{equation}
    d = \frac{1}{|f|}\sum_{f} {H^{f}[\tilde{d}^{f^c}]},
\end{equation}
where $|f|$ denotes the number of adjacent triangles, $H^f$ is the local coordinate of the triangle $f$ in the coarse garment $C$. 
Finally, we compute the corrected low-resolution garment mesh $\hat{C}$ by adding the displacement $D$ to the coarse garment input $C$, that is: $\hat{C}=C+D$.

\subsection{Detail Wrinkle Residual Using  $\mathcal{A}$} \label{subsec: net A}
We locally construct the continuous up-sampling detail wrinkle field $\mathcal{W}^f$ for each coarse garment mesh triangle $f$. Employing a hyper-network $\mathcal{A}$, we decode the parameters $\{w\}^f$ of the implicit function $\mathcal{W}^f$ using the detail super-resolution features $\tilde{z}^{f^0}$, $\tilde{z}^{f^1}$, $\tilde{z}^{f^2}$ corresponding to the three vertices of the coarse mesh triangle $f$:
\begin{equation}
    \{w\}^f = \mathcal{A}[\tilde{z}^{f^0}, \tilde{z}^{f^1}, \tilde{z}^{f^2}],
\end{equation}
where $\mathcal{A}$ is an MLP network.
Each up-sample vertex $k$ of the high-resolution mesh possesses a barycentric coordinate $(u^k,v^k)$ with respect to its corresponding triangle $f$ in the coarse mesh topology.
We apply the constructed $\mathcal{W}^f$ to decode the fine detail wrinkle residual $r^k$ with the barycentric coordinate $(u^k,v^k)$ as input:
\begin{equation}
    r^k = \mathcal{W}^f[1-u^k-v^k, u^k, v^k].
\end{equation}
Then, we compute the position of any high-resolution garment mesh vertex $g^k$ as:
\begin{equation}
    g^k = \mathcal{S}[\hat{C},f,u^k, v^k] + \hat{H}^f[r^k],
\end{equation}
where $\hat{C}$ is the corrected coarse garment shape as described in Section \ref{subsec: net C}, and $\hat{H}^f$ denotes the local coordinate of the triangle $f$ in $\hat{C}$. 

Specifically, $\mathcal{W}^f$ is implemented as a complex MLP. We configure the number of layers in $\mathcal{W}^f$ to be $5$, with perceptrons $x$ activated by the complex Gabor wavelet activation function $\mathbf{Wire}(x, \omega_0, \alpha_0)$ except for the last layer of output.
This choice exhibits superior performance in high-frequency feature reconstruction. We set the frequency control $\omega_0$ and the spread control $\alpha_0$ to be $5$ and $10$, respectively. For additional insights, please refer to \cite{saragadam2023wire}. 

\subsection{Loss Function}
As L1 loss demonstrates superior performance in super resolution tasks \cite{zhao2017loss}, we train our \emph{GDSR} using objective L1 loss functions to ensure the accuracy of both coarse garment correction and wrinkle detail generation.

\paragraph{Coarse garment correction.}
We first consider L1 loss with respect to the corrected low-resolution garment geometry $\hat{C}_t$ and its normal $\hat{N}_t$ to enforce a similarity to the ground truth $\hat{C}_t^*$, as follows:
\begin{equation}
    L_{c\_geo}(\hat{C}_t) = \|\hat{C}_t - \hat{C}_t^*\|_1 + 0.01 \|\hat{N}_t-\hat{N}_t^*\|_1,
\end{equation}
where $\hat{C}_t^*$ is predefined before training by down-sample the ground truth of high-resolution garment mesh $G_t$, and $\hat{N}_t^*$ denotes the normal of $\hat{C}_t^*$.
Additionally, we enforce the stretching and shearing of $\hat{C}_t$ to be similar as the ground truth $\hat{C}_t^*$. We approximate the mapping matrix:
\begin{equation}
    w(\hat{C}_t):=\begin{pmatrix} w_u(\hat{C}_t) | w_v(\hat{C}_t) \end{pmatrix}_{3\times 2},
\end{equation}
over all triangles from the 2D cloth plane coordinate $(u, v)$ to the garment $\hat{C}_t$ in the world space. For more insights of the mapping matrix $w(\hat{C}_t)$, please refer to \cite{kim2020dynamic, baraff2023large}. Thus, we define the objective loss function as:
\begin{align}
    L_{c\_def}(\hat{C}_t) = & 10^4 \|\Delta \hat{C}_t\|^2 + 0.1\|\theta(\hat{C}_t) - \theta(\hat{C}^*_t)\| \\ 
    & + \|\|w_u(\hat{C}_t)\| - \|w_u(\hat{C}^*_t)\|\|_1 \\
    & + \|\|w_v(\hat{C}_t)\| - \|w_v(\hat{C}^*_t)\|\|_1 \\ 
    & + \|w_u(\hat{C}_t)\cdot w_v(\hat{C}_t) - w_u(\hat{C}^*_t)\cdot w_v(\hat{C}^*_t)\|_1,
\end{align}
where $\Delta$ is the Laplacian operator and $\theta$ denotes dihedral angles between adjacent mesh triangles of $\hat{C}$.
The final loss function for correcting the low-resolution garment mesh is a linear combination of the terms:
\begin{equation}
    L_{coa}({\hat{C}_t}) = L_{c\_geo}(\hat{C}_t) + 0.01 L_{c\_def}(\hat{C}_t).
\end{equation}
%
%
\begin{wrapfigure}{R}{0.28\columnwidth}
\centering
\includegraphics[width=0.28\columnwidth]{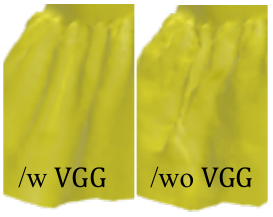}
\end{wrapfigure}


\paragraph{Patch-based learning wrinkle detail generation.} 
We adopt a patch-based strategy to learn wrinkle detail generation. 
The high-resolution patches, denoted as $P$, are cropped based on the garment's 2D UV texture image, which maintains a consistent unit scale between pixels and 3D space, ensuring each mesh triangle corresponds to at least one pixel. 
As the normal map $\mathcal{N}$ sensitively describes wrinkle details, we render the normal map for both the predicted patch $P_t$ and the ground truth patch $P_t^*$ of high-resolution garment simulation. We define a reconstruction loss concerning vertex positions and multi-layer features of the pre-trained $\mathbf{VGG}$ network {\cite{Simonyan15}}:
\begin{equation}
    L_{w\_geo}(P_t) = \|P_t - P_t^*\|_1 + 0.01 \sum_\ell \|\mathbf{VGG}^\ell[\mathcal{N}^P_t]-\mathbf{VGG}^\ell[\mathcal{N}^{P^*}_t]\|_1.
\end{equation}
As seen in the inset figure, utilizing $\mathbf{VGG}$ features (/w $\mathbf{VGG}$) helps capture more reasonable high-frequency wrinkle patterns, in comparison to the result without using $\mathbf{VGG}$ (/wo $\mathbf{VGG}$).
To constraint the deformation of each triangle on the wrinkle detailed patches, we derive $w_u(P_t)$ and $ w_v(P_t)$ from the estimation of mapping matrix $w(P_t)$ to define the deformation loss as:
\begin{align}
    L_{w\_def}(P_t) = & \|\Delta P_t\|^2 + \|\|w_u(P_t\| - \|w_u(P_t^*)\|\|_1 \\ 
    & + \|\|w_v(P_t)\| - \|w_v(P_t^*)\|\|_1 \\ 
    & + \|w_u(P_t)\cdot w_v(P_t) - w_u(P_t^*)\cdot w_v(P_t^*)\|_1.
\end{align}
The final loss function for the wrinkle detail generation is defined as:
\begin{equation}
    L_{wri} = \frac{1}{|P|}\sum_{P}L_{w\_geo}(P_t) + 0.01 L_{w\_def}(P_t),
\end{equation}
where $|P|$ is the number of sampled patches.
\paragraph{Training.}
We jointly train the three network blocks in \emph{GDSR}, which consist of {Mesh-Graph-Net} $\mathcal{H}$, {Decoder} $\mathcal{E}$ and {Mesh-Hyper-Net} $\mathcal{A}$, to enhance high-resolution wrinkle details while correcting the low-resolution garment geometry. Therefore, we define the final loss function for \emph{GDSR} as:
\begin{equation}
    L_{GDSR} = \lambda_c L_{coa}(\hat{C}_t) + \lambda_w L_{wri}(\{P_t\}).
\end{equation}
Initially, we set $\lambda_c=1.9$ and $\lambda_w=0.1$ since coarse garment correction is a prerequisite for wrinkle detail generation. 
{During training, $\lambda_c$ is decreased and $\lambda_w$ is increased by $0.2$ every $10$ epochs until both reach $1$.}
We then continue training \emph{GDSR} with $\lambda_c=1$ and $\lambda_w=1$ until the loss converges. 
{We use AdamW Optimizer with an initial learning rate of $1 \times 10^{-4}$, halving it every $50$ epochs. Convergence typically occurs after around $300$ epochs. Due to the varied mesh topology of the different garment types in our training data, we set the batch size to $1$ for computing $L_{coa}$, while randomly cropping $8$ normal patches from the high-resolution mesh for $L_{wri}$.}


\subsection{Explicit Collision Handling}
\begin{figure}[t]
   \centering
    \includegraphics[width=0.68\columnwidth]{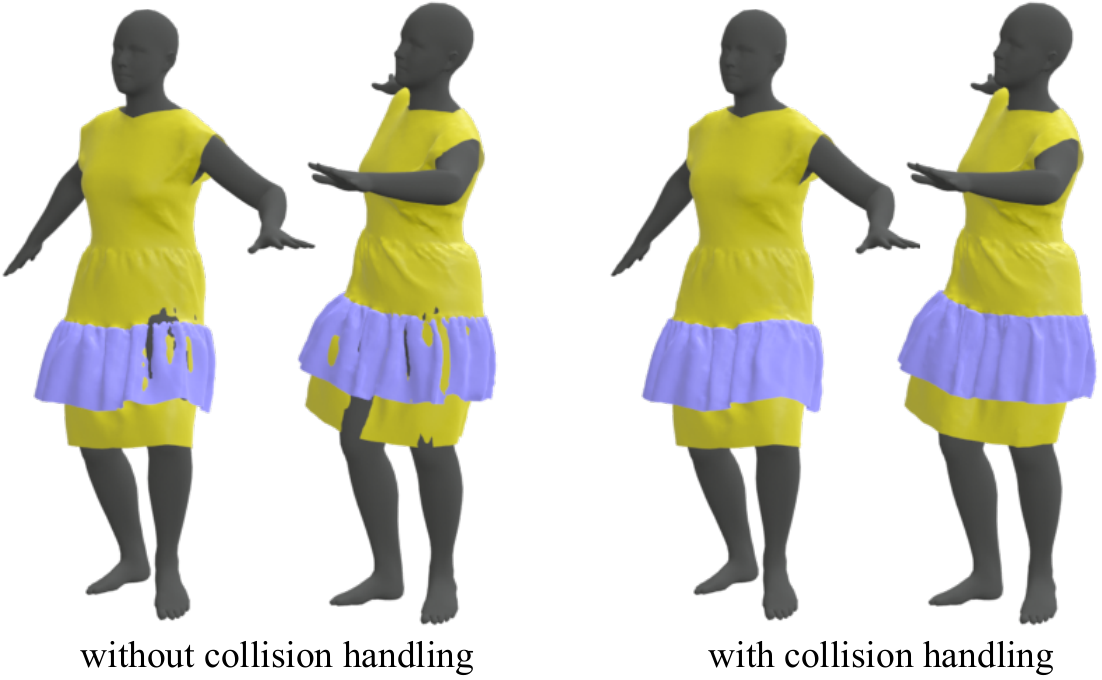}
    \vskip -0.15in
    \caption{\textbf{Explicit collision handling.} We detect the garment collisions with the body and that between the fabric layers at a coarse level of garment geometry, and propagate the detection to efficiently resolve the collisions for high resolution garment.}
    \label{fig:collisions}
\end{figure}

We observe that garment collisions with the body and between different fabric layers are inevitable, especially in the case of unseen garments and body motions. Efficient collision handling alongside high-resolution garment dynamics is too expensive for constrained budgets. To address this, we detect potential collisions on the low-resolution mesh and propagate this information to the high-resolution garment. 
When running \emph{GDSR}, we identify potential collisions between the corrected coarse garment $\hat{C}_t$ and the body $B_t$. For a coarse garment vertex $\hat{c}_t^i$, a potential collision with a body vertex $b_t$ is identified if $\sigma_b - (\hat{c}_t^i - b_t) \cdot n_t^b > 0$, where $\sigma_b$ is a predefined constant, and $n_t^b$ is the tangent plane normal vector of $b_t$. Subsequently, a high-resolution garment vertex $g_t^k$, where $\hat{c}_i$ is the nearest low-resolution garment vertex to $g^k$ in the 2D UV parameterization space, is adjusted to a new position:
\begin{equation}
g^k_t = g^k_t + \max[\sigma_b -(g^k_t - b_t) \cdot n_t^b, 0]n_t^b.
\end{equation}
For a coarse garment vertex $\hat{c}_t^{\ell_e}$ in the exterior layer, we detect a collision with an interior layered garment vertex $\hat{c}t^{\ell_i}$ if $\sigma\ell - (\hat{c}_t^{\ell_e} - \hat{c}_t^{\ell_i}) \cdot n_t^{\ell_i} > 0$ and $|\hat{c}_t^{\ell_e} - \hat{c}t^{\ell_i}| < \gamma$, as the garment mesh is an open surface. Here $\sigma\ell$ and $\gamma$ are predefined constants. We resolve the collision between fabric layers of a high-resolution garment by adjusting the exterior layer vertex $g_t^{\ell_e}$ to account for the collision of the nearest coarse garment vertex $\hat{c}_t^{\ell_e}$ with the interior garment vertex $\hat{c}_t^{\ell_i}$:
\begin{equation}
    g_t^{\ell_e} = g_t^{\ell_e} + \max[\sigma_\ell - (g_t^{\ell_e}-\hat{c}_t^{\ell_i}) \cdot n_t^{\ell_i}, 0]n_t^{\ell_i}.
\end{equation}
Finally, we resolve garment collisions with the body and between fabric layers by updating the high-resolution garment $G_t$. The effect of this explicit collision handling is shown in Figure \ref{fig:collisions}.

\begin{figure*}[!t]
    \centering
    \includegraphics[width=0.87\textwidth]{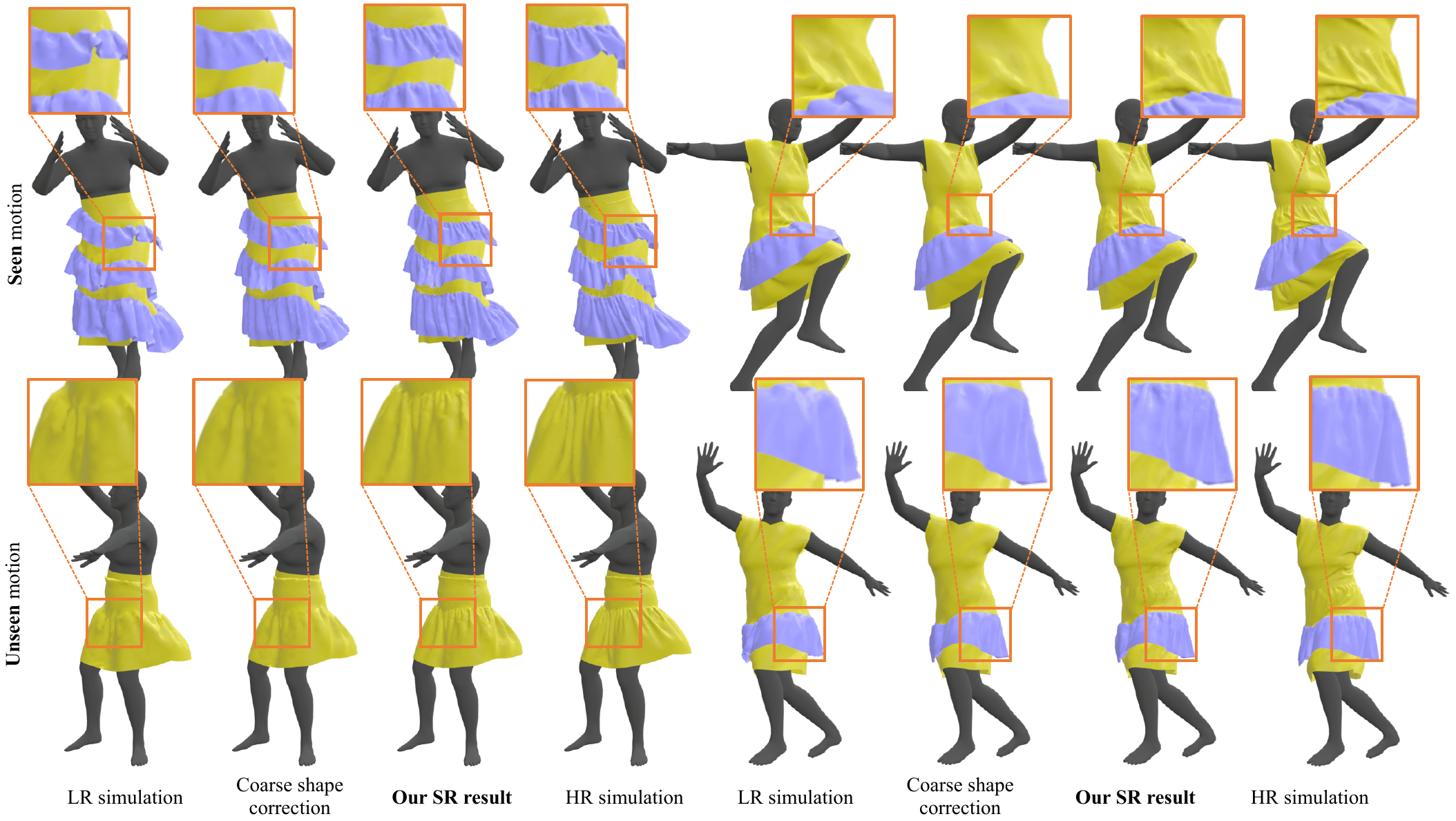}
    \vskip -0.12in
    \caption{ We run our method by roll-out prediction on both seen (1st row) and unseen (2nd row) motions with the training garment types. For each example, we show input of low resolution simulation, results of coarse shape correction and detailed high resolution results,  as well as the reference of high resolution simulation.}
    \label{fig:seen_garments}
\end{figure*}

\begin{figure*}[!t]
    \centering
    \includegraphics[width=0.90\textwidth]{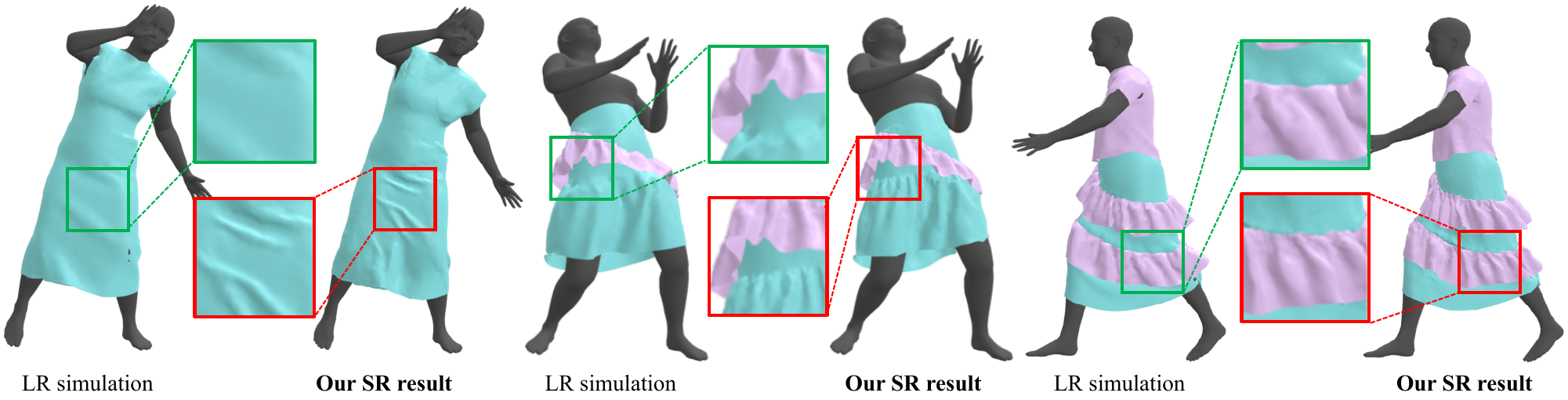}
    \vskip -0.12in
    \caption{We evaluate the generalization of our method on motions and garment types that are both out of the training data. Our method can synthesize fine-grained wrinkle details for either single- or multi-layer garments.}
    \label{fig:unseen_garments}
\end{figure*}

\begin{figure*}[!t]
   \centering
   \includegraphics[width=0.90\textwidth]{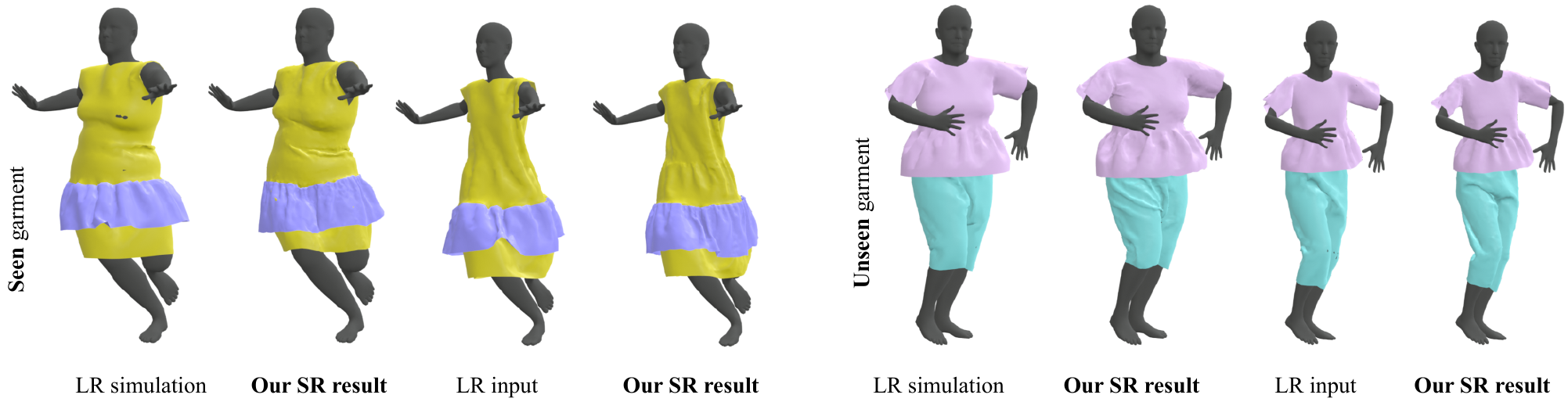}
       \vskip -0.12in
   \caption{
    We evaluate the generalization of our method across different body shapes on out-of-training motions with both seen and unseen garment types. 
    Although trained with a fixed body shape, our method can synthesize detailed high-resolution garment geometry fitting with varying body shapes. 
  }
   \label{fig:bodyshape}
 \end{figure*}
 
\section{Experiments and Results} \label{sec: results}

Using two dancing sequences (swing and house) on five garment outfits (colored yellow or purple) with a fixed body shape, we train \emph{GDSR} network with garment dynamic history states, derived from high-resolution garment ground truth. The first row in Figure 5 shows the results of our method testing on seen motions with the training garment types in a roll-out prediction manner. During iterative processing on the previous predictions and low-resolution simulation inputs, our method robustly produces fine-grained garment wrinkle details resembling high-resolution simulation references.

To test our method's generalization ability, we generate seven additional motion sequences and five garment outfits (colored teal or pink). The differences between the training and testing data are illustrated in the supplementary material. Our method successfully generalizes across various body motions, garment types, and body shapes not present in the training data. Sampled frames of our results are shown here. For a better evaluation of high-resolution garment details, please refer to the supplementary video.

\paragraph{Generalization to unseen motions.}

Instead of using body motions as global signals to drive garment dynamics, our network focuses on local garment dynamics and interactions with the underlying body to learn local displacement vectors and wrinkle detail residuals. The second row in Figure \ref{fig:seen_garments} demonstrates our method's generalization ability to unseen motion sequences (samba dancing and samba variant). Our method synthesizes fine-grained wrinkle details, producing wrinkle patterns similar to those in high-resolution simulations.


 \paragraph{Generalization to unseen garments.}
Based on our training data, which includes multiple garment types, we generate new garments by removing or adding lace embellishments, modifying garment lengths, or combining different garments. Our method achieves remarkable generalization by using Mesh-Graph-Net, which is not restricted to any specific mesh topology, and Hyper-Net to construct continuous up-sampling fields for any local mesh triangle. 
Figure \ref{fig:unseen_garments} showcases the results of our method on body motions and garment types that are not part of the training data. Our results demonstrate that our method can synthesize fine-grained wrinkle details for both single- and multi-layered garments.


 \paragraph{Generalization to different body shapes.}
 Our method is trained using a fixed body shape. In Figure \ref{fig:bodyshape}, we first test its generalization to different body shapes with a garment in training data. The varying wrinkle patterns on the garment for different body shapes demonstrate that our method can synthesize detailed, high-resolution garment geometry that fits varying body shapes. Next, we design a combination of a lace t-shirt and pants in two different sizes to fit two body shapes. The results in Figure \ref{fig:bodyshape} show that our method performs plausibly on unseen garments with different body shapes.
 
 \begin{figure}[!htpb]
   \centering
   \includegraphics[width=\columnwidth]{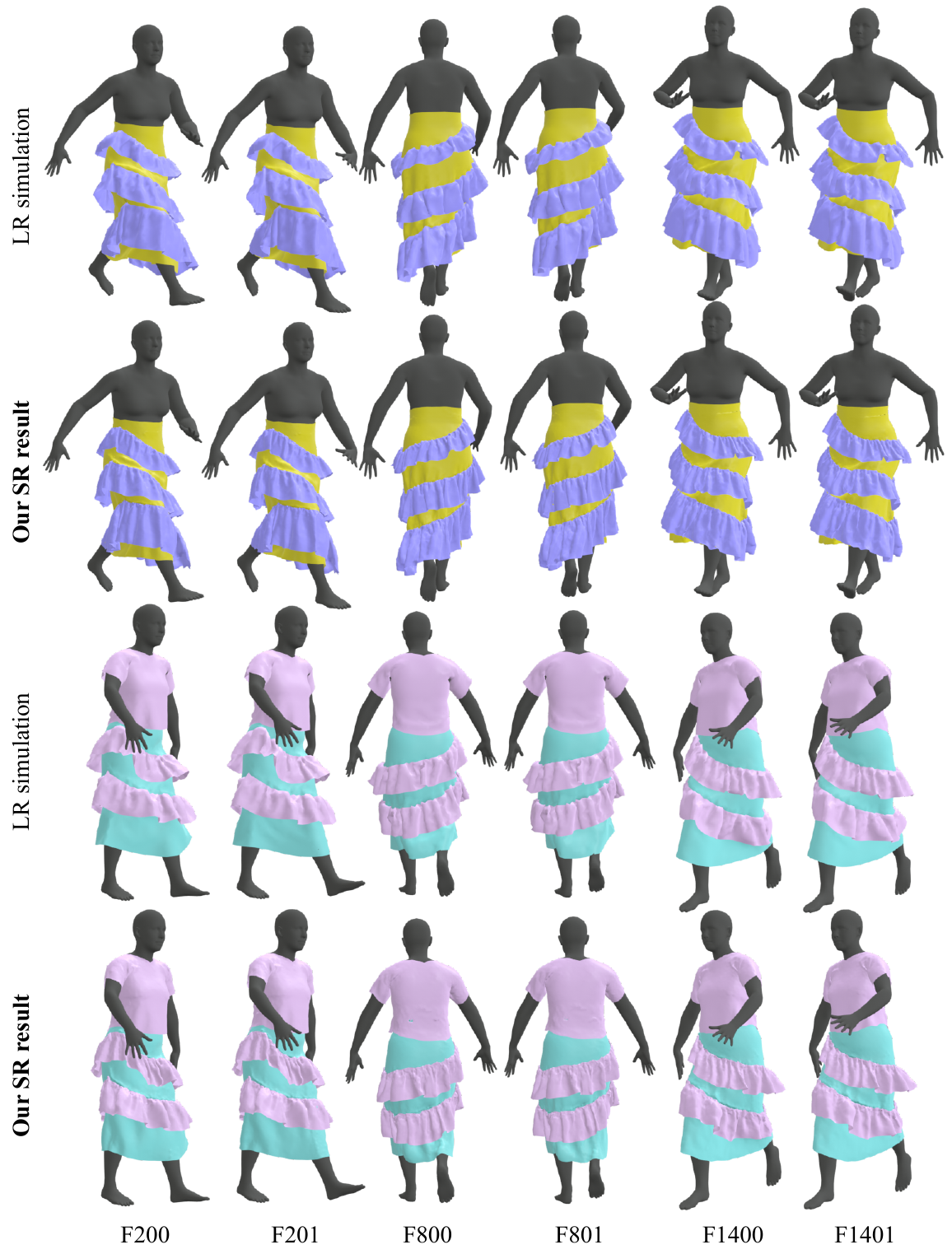}
       \vskip -0.1in
   \caption{
    We evaluate the robustness of our method under long roll-out predictions up to 1500 frames, using a seen garment (in purple and yellow) and an unseen garment (in teal and pink) with happy walking and normal walking motion sequences, respectively. Our method uses low-resolution simulation and its predictions as input for subsequent frames to produce reasonable wrinkle details while maintaining plausible temporal smoothness between frames.
  }
   \label{fig:longrollout}
 \end{figure}

 \begin{figure}[!htpb]
   \centering
    \includegraphics[width=0.95\columnwidth]{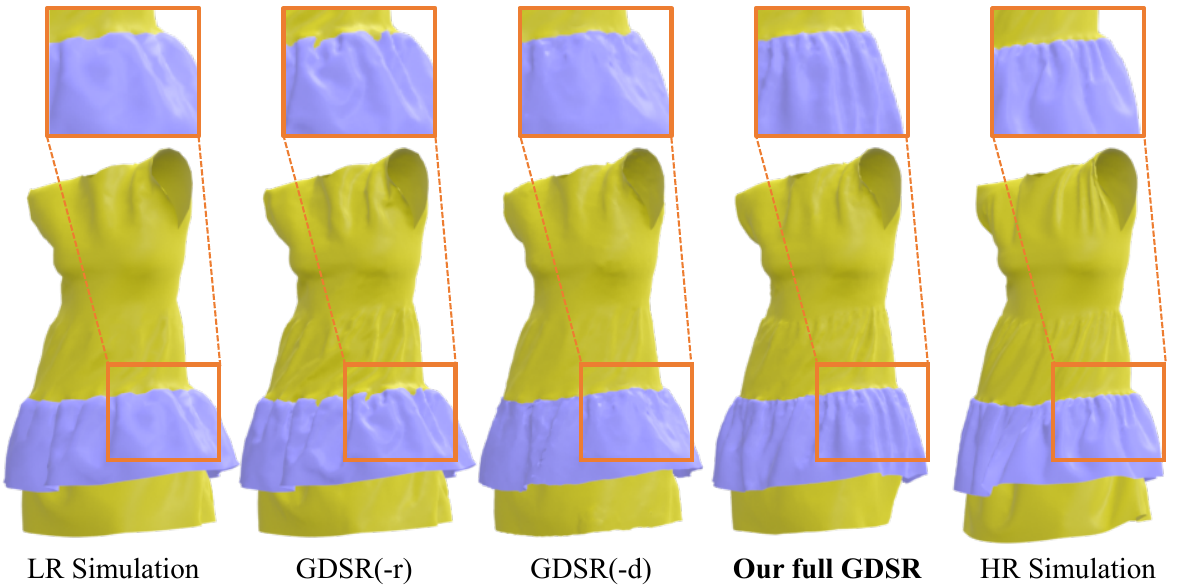}
    \vskip -0.1in
    \caption{
    To evaluate the effect of coarse garment correction, we designed two variants of GDSR by removing the \emph{Decoder} $\mathcal{E}$, named GDSR(-r) and GDSR(-d) respectively. Our full model of GDSR produces remarkably fine-grained high-resolution garment geometry with wrinkle detail patterns similar to the reference of high-resolution simulation, compared to GDSR(-r), which does not consider the influence of coarse garment shape, and GDSR(-d), which relies on roll-out prediction but without correcting the garment shape at coarse level.}
    \label{fig:coarse_correction}
\end{figure}
 
 \paragraph{Robustness of iterative roll-out prediction.}
 Our method is trained with two separate dance motion sequences: \emph{House} (714 frames) and \emph{Swing} (794 frames). We evaluate the robustness of iterative roll-out prediction on walking sequences up to 1500 frames, testing on both seen and unseen garment types. As shown in Figure \ref{fig:longrollout}, our network uses its predictions as input for subsequent frames, producing reasonable predictions for long testing sequences. The wrinkle details synthesized by our method maintain plausible structural coherence between consecutive frames. We evaluate the stability of long roll-out prediction quantitatively in the supplemental material.


 \paragraph{Effect of coarse shape correction.}
 To evaluate the effect of coarse garment shape correction, we design two variants of \emph{GDSR} by removing the \emph{Decoder} $\mathcal{E}$, named GDSR(-r) and GDSR(-d) respectively. Without considering the influence of coarse garment shape on the dynamics of wrinkle details, we train GDSR(-r) using a down-sampled coarse garment $\hat{C}$ from the high-resolution simulation as input. This is done by initializing the input of node features with vectors of normal, velocity, and acceleration computed on the down-sampled geometry $\hat{C}$, and setting the displacements $d_{t-1}$ to 0. As shown in Figure \ref{fig:coarse_correction}, with low-resolution simulation as input, GDSR(-r) only emphasizes the original wrinkle details present on the coarse garment geometry. Meanwhile, GDSR(-d) involves the history of high-resolution detail predictions in its subsequent predictions but, without the {Decoder} $\mathcal{E}$ to correct the coarse garment shape at the coarse level, it fails to capture clear high-frequency details. In comparison, our full model of \emph{GDSR} excels in capturing fine-grained wrinkle detail patterns similar to the reference of high-resolution simulation with coarse garment shape correction.

\begin{figure}[!t]
   \centering
    \includegraphics[width=1\columnwidth]{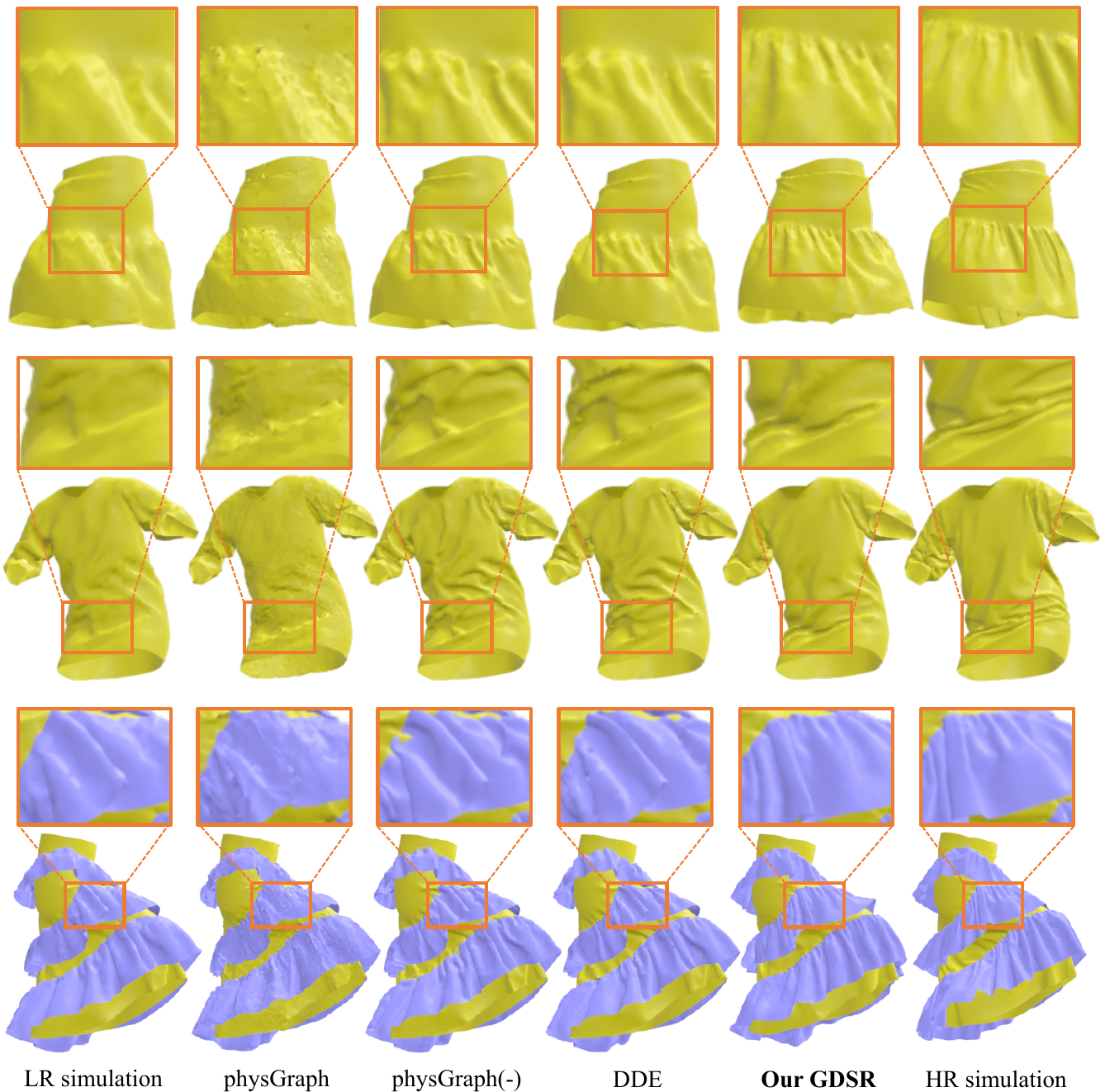}
    \vskip -0.1in
    \caption{Given the low-resolution simulation garment geometry, our GDSR constructs the high-resolution garment geometry by synthesizing fine detail wrinkles, applied on the corrected coarse garment mesh. Comparing to PhysGraph \cite{halimi2023physgraph} and DDE\cite{zhang2021deep}, GDSR captures fine-grained wrinkle patterns more similar to the reference high-resolution simulation.}
    \label{fig:comparisons}
\end{figure}

\subsection{Comparisons}
We evaluate our approach by comparing it with two recent techniques for enhancing garment details: DDE \cite{zhang2021deep} and PhysGraph \cite{halimi2023physgraph}. DDE is an image-based method that enhances wrinkles by applying style transfer to normal images. We train DDE using downsampled garment geometry from high-resolution simulations. However, some high-frequency details are not captured because DDE does not consider the impact of coarse garment shape on garment wrinkles, and some details are smoothed out by the subsequent normal-guided deformation.
PhysGraph uses a mesh-based graph network but operates on high-resolution garment meshes up-sampled from low-resolution simulations. PhysGraph fails to converge effectively, resulting in noisy garment meshes when trained with the same data as \emph{GDSR}. To address this, we train a variant (PhysGraph(-)) using down-sampled garment geometry from high-resolution simulations, similar to DDE. Without correcting the coarse garment shape, PhysGraph(-) achieves only comparable quality to DDE. In contrast, our \emph{GDSR} captures fine wrinkle details that closely resemble the high-resolution simulation reference (see Figure \ref{fig:comparisons}). We report the quantitative comparisons with style improvement score in the supplemental material.

\subsection{Running Efficiency} \label{subsec: runningtime}
Our lightweight network, with a storage size of 65MB, is potentially compatible with consumer devices like smartphones. 
{Despite unoptimized PyTorch code and low-resolution simulation efficiency, we believe our method has the potential to enable real-time, high-resolution clothing simulations. For instance, it takes only 0.115 seconds to synthesize one frame of detailed high-resolution garments with 43,257 vertices from coarse geometry with 5,010 vertices, including 0.083 seconds for low-resolution simulation in Marvelous Designer (MD). The supplemental material provides more details on computation time evaluation of \emph{GDSR} and comparisons with physics-based simulations in MD and two baseline methods \cite{zhang2021deep, halimi2023physgraph}.}

\section{LIMITATIONS \& FUTURE WORK}

\begin{figure}[!h]
   \centering
    \includegraphics[width=\columnwidth]{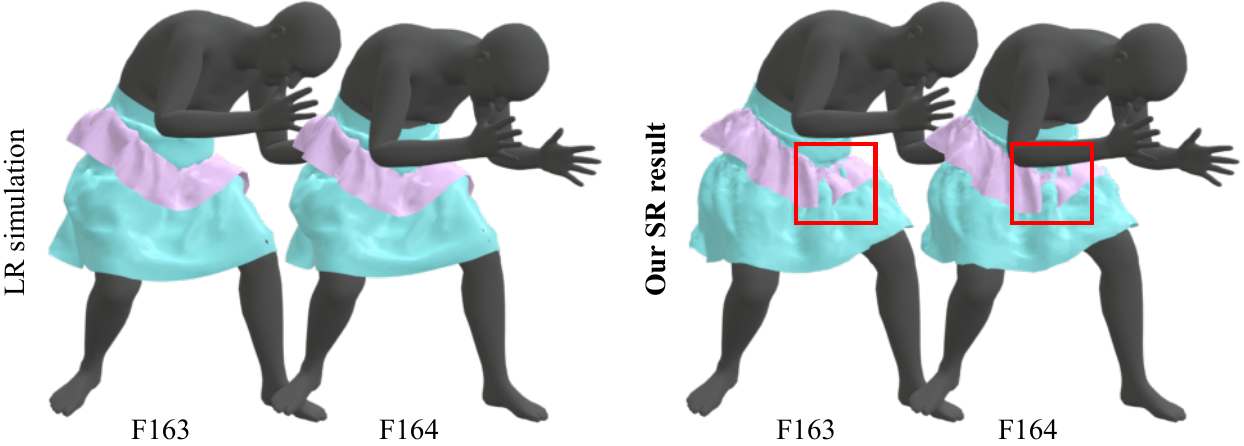}
    \vskip -0.1in
    \caption{\textbf{Limitation.} Our method fails to capture garment details influenced by friction between fabric layers. Although our explicit collision handling resolves most intersections between fabric layers, some collisions remain because we cannot detect intersections when the distances between intersecting vertices exceed the predefined threshold.}
    \label{fig:limitation}
\end{figure}

Our work has several limitations. As shown in Figure \ref{fig:limitation}, \emph{GDSR} fails to capture wrinkle details influenced by fabric friction, as it lacks graph features to describe this friction. Additionally, due to low-resolution collision detection, some collisions remain unresolved, especially in multi-layered garments with intersecting vertices beyond the predefined threshold. Existing works \cite{santesteban2022ulnef, lee2023clothcombo} untangle the intersection of multi-layered garments specific for static body poses but not for dynamic motions. Future research could explore collision resolution for dynamic multi-layered garments using signed distance fields for local meshes.
Furthermore, we aim to extend garment dynamic super-resolution to integrate wrinkle synthesis and garment appearance rendering, achieving better structural consistency across time and space than previous works \cite{10.1145/3478513.3480497}. Finally, as \emph{GDSR} is trained on a single fabric material, we could introduce conditioned instance normalization to adjust latent features for different materials \cite{zhang2021deep}, or explore meta-learning for high-resolution garment simulations to meet the apparel industry's customization needs.

\begin{acks}
We would like to thank the anonymous reviewers for their
constructive comments; 
Kaizhang Kang for his generous helping with text, graphics, and video polishing;
Mixamo for the motion sequences.
This work was partially supported by the National Science Fund of China (No. 62472223 and No. 62072242). 
\end{acks}

\bibliographystyle{ACM-Reference-Format}
\bibliography{main_bib}

\appendix
\begin{figure}[h!]
    \centering
     \includegraphics[width=0.8\columnwidth]{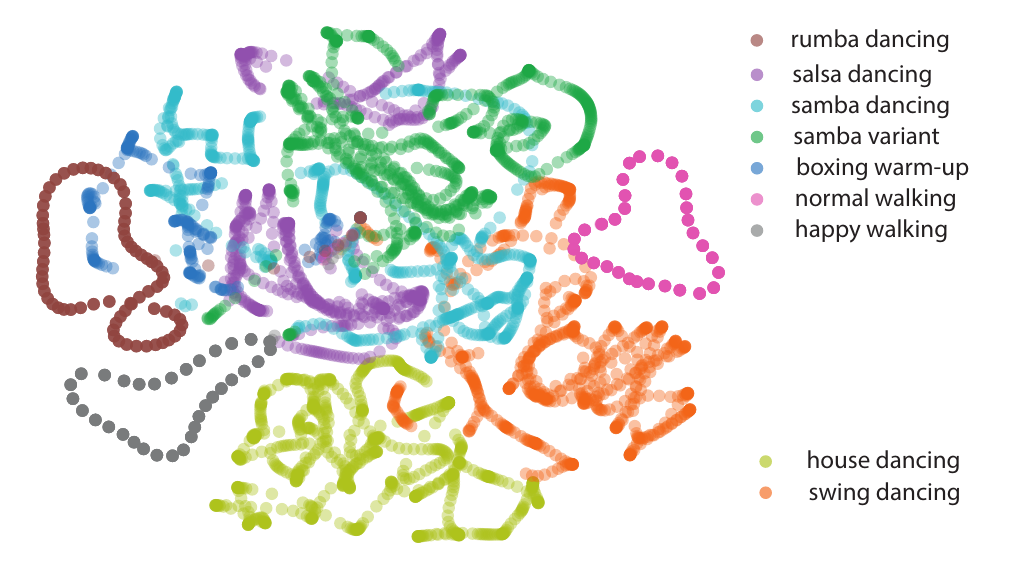}
    \caption{We visualize the distribution of motion sequences used for generating training data (House and swing) and the testing data (rumba, salsa, samba, samba variant, boxing warm-up, normal walking, happy walking) via t-SNE \cite{van2008visualizing}.}
    \label{fig:tSNE}
\end{figure}


\begin{figure}[h!]
    \centering
     \includegraphics[width=0.85\columnwidth]{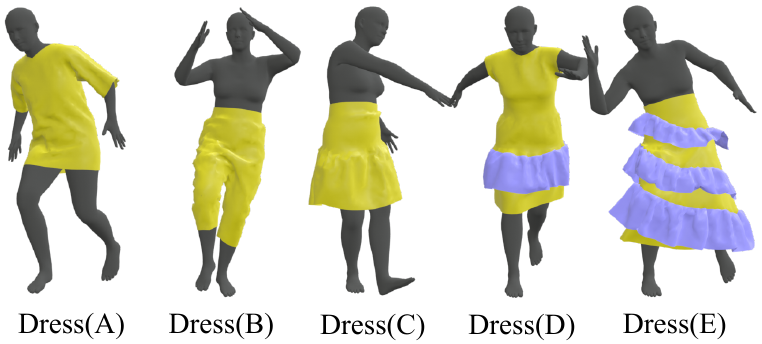}
    \caption{We design 5 garment outfits to train the network of our method.}
    \label{fig:train_garment}
\end{figure}


\section{Data Generation}
We employ SMPL \cite{10.1145/2816795.2818013} to extract body shapes, and subsequently utilize rigging and animation techniques through Mixamo\footnote{https://www.mixamo.com/} to generate $9$ sequences of motion. 

We train our network using dance sequences of \emph{swing dancing} {($714$ frames)} and \emph{house dancing} {($794$ frames)} on a fixed body shape, while we conduct testing on rumba dancing (365 frames),{salsa dancing} {($580$ frames)}, {samba dancing} {($609$ frames)}, samba variant ($513$ frames), {boxing warm-up} {($203$ frames)}, {{normal walking} (1510 frames)} and {{happy walking} (1510 frames)}. We visualize the distribution of the motion sequences via t-SNE \cite{van2008visualizing} in Figure \ref{fig:tSNE}. 

To generate garments, we employ physics-based simulation through Marvelous Designer\footnote{https://marvelousdesigner.com/} with a material of silk chamuse and generate low- and high-simulation with particle distance set at $30$mm and $10$mm, respectively. We model $5$ garment outfits (colored in yellow or purple) to generate training data with the pre-defined training motion sequences: (A) a long sleeve t-shirt, (B) a pair of long pants, (C) a pleated short skirt, (D) a double-layered dress, (E) a triple-lace long skirt. We illustrate the garment types of training data in Figure \ref{fig:train_garment}.

\begin{figure}[t]
    \centering
     \includegraphics[width=0.85\columnwidth]{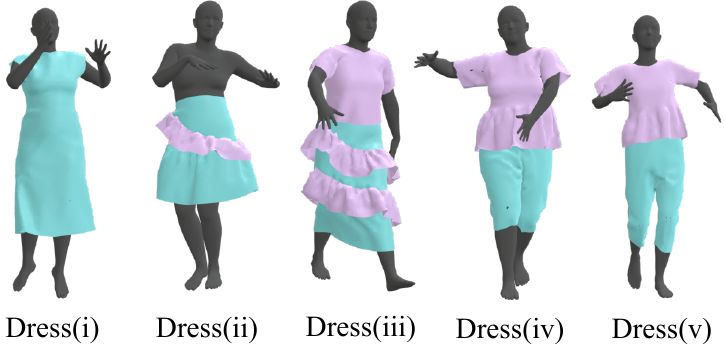}
    \caption{We design 5 garment outfits to test the generalization ability of our method across different garment types.}
    \label{fig:test_garment}
\end{figure}
To test the generalization capability of our method across out-of-training garment types, we generate $5$ new garment outfits (colored in teal or pink): (i) a single-layered long dress, (ii) a laced pleated short skirt, (iii) a full-body combination of short sleeve t-shirt and double-lace long skirt, and a full-body combination of lace t-shirt and pants in large size (iv) and small size (v). We illustrate the garment types of training data in Figure \ref{fig:test_garment}.
With the low-resolution simulation as input, we train our network in a fully supervised manner with the high-resolution simulation as the ground truth. We test our network directly with the low resolution simulation as input. We list the number of garment vertices of both our training and testing data in Table \ref{tab:garment_vertices}.

\begin{table}[h!]
    \caption{We list the number of garment vertices for both our training and testing data. Our method generates high resolution garment geometry with approximately 10 times the number of vertices of the low resolution simulation input.}
    \begin{tabular}{c|c|c||c|c|c}
    \multicolumn{3}{c||}{training garment} & \multicolumn{3}{c}{testing garment} \\
    \hline
   Dress & LR & HR & Dress & LR & HR \\
        \hline
    (A) & 2063 & 18445 & (i) & 2485 & 22044 \\
    \hline
    (B) & 1783 & 16083 & (ii) & 2272 & 19618 \\
    \hline
    (C) & 1161 & 14528 & (iii) & 4771 & 41554 \\
    \hline
    (D) & 3176 & 27855 & (iv) & 3607 & 32025 \\
    \hline
    (E) & 5010 & 43257 & (v) & 2519 & 22538 \\
    \hline
    \end{tabular}
    \label{tab:garment_vertices}
\end{table}


\section{Robustness of long roll-out prediction}
We use average stretching and shearing energies\cite{10.1145/280814.280821} of garment mesh triangles to evaluate the stability of high-resolution garment deformation with our method under long roll-out predictions. We compute the deformation mapping matrix $w(G_t)$ from the 2D cloth plane coordinate $(u, v)$ to the garment $G_t$ in the world space. We define the stretching energy as $E_u = \|\|w_u(G_t)\|-1\|^2$ and $E_v = \|\|w_v(G_t)\|-1\|^2$, along the $u$ and $v$ directions respectively. Thus, the garment mesh triangle is un-stretched whenever $E_u = 0$ and $E_v = 0$. 
We define the shearing energy as $E_{uv} = (w_u(G_t)\cdot w_v(G_t))^2$.  When $E_{uv} = 0$, the triangle is un-sheared. 
Table \ref{tab:rollout} shows that under long roll-out predictions of up to 1500 frames, our method produces reasonable high-resolution garment geometry. The stretching and shearing energies remain stable within a small range, preventing excessive deformations.


\begin{table}[h!]
    \caption{
    We apply the average stretching energy ($E_u$, $E_v$) and shearing energy ($E_{uv}$) of the garment mesh triangles to evaluate the stability of garment mesh deformation with our method under long roll-out predictions. Our method produces reasonable high-resolution garment geometry with the stretching and shearing energies remaining stable within a small range, preventing excessive deformations.
    }
    \begin{tabular}{c||c|c|c|c|c}
        \hline
        \multicolumn{6}{c}{seen garment: Dress(E)}  \\
        \hline
        & 1-step & roll-50 & roll-100 & roll-1000 & roll-1500\\
         \hline
        $E_u$ & 0.020 & 0.038 & 0.029 & 0.028 & 0.029 \\
        \hline
        $E_v$ & 0.012 & 0.022 & 0.017 & 0.013 & 0.016 \\
        \hline
        $E_{uv}$ & 0.034 & 0.050 & 0.048 & 0.041 & 0.046 \\
        \hline
        \multicolumn{6}{c}{unseen garment: Dress(iii)}  \\
         \hline
         & 1-step & roll-50 & roll-100 & roll-1000 & roll-1500\\
         \hline
        $E_u$ & 0.028 & 0.031 & 0.032 & 0.041 & 0.036 \\
        \hline
        $E_v$ & 0.012 & 0.016 & 0.019 & 0.024 & 0.019 \\
        \hline
        $E_{uv}$ & 0.042 & 0.056 & 0.056 & 0.065 & 0.059 \\
        \hline
    \end{tabular}
    \label{tab:rollout}
\end{table}

\section{Comparisons}
{
\begin{table}[h!]
    \caption{
    With low-resolution simulation as input, we employ the Structural Similarity Index Measure (SSIM) \cite{wang2004image} to compare the performance of our GDSR, DDE\cite{zhang2021deep}, PhysGraph \cite{halimi2023physgraph} and its variant (PhysGraph(-)) on a t-shirt (A), a pleated short skirt (C), and a triple-lace long skirt (E). We report the average SSIM over 300 frames of normal maps from an unseen motion sequence. 
    }
    \begin{tabular}{c||c|c|c|c}
   Dress & PhysGraph & PhysGraph(-) & DDE & GDSR \\
    \hline
    (A) & 0.831 & 0.848 & 0.855 & \textbf{0.879} \\
    \hline
    (C) & 0.746 & 0.779 & 0.783 & \textbf{0.816}\\
    \hline
    (E) & 0.685 & 0.723 & 0.735 & \textbf{0.788} \\
    \hline
    \end{tabular}
    \label{tab:ssim_comparison}
\end{table}

To evaluate the similarity of the detailed wrinkle patterns to the high-resolution simulation, we conducted a quantitative assessment using the Structural Similarity Index Measure (SSIM) \cite{wang2004image} on sequences of normal maps rendered from the garment geometries. As shown in Table \ref{tab:ssim_comparison}, our method outperforms the baseline approaches \cite{zhang2021deep, halimi2023physgraph}, producing high-resolution garment geometries with wrinkle patterns that more closely align with the high-resolution simulation references in terms of wrinkle detail structure.
}

\section{Computation time}
Table \ref{tab:time-cost} presents the computation time required to simulate a single frame of high-resolution garment geometry for Dress (C), (D), and (E), with 14,528, 27,855, and 43,257 vertices respectively. Compared to the high-resolution physics-based simulation in Marvelous Designer (MD), our method achieves significantly better efficiency, particularly for complex garment types. For instance, while the high-resolution simulation with MD takes 0.559 seconds, our GDSR method requires only 0.115 seconds to enhance wrinkle details on the low-resolution simulation, including 0.083 second running in MD for coarse garment generation.

We also compared the computation time of our method with DDE and PhysGraph. Our method synthesizes high-resolution wrinkle details on the coarse garment shape for Dress (C) in just 0.014 seconds, significantly outperforming DDE, which takes 0.202 seconds due to its time-consuming normal-guided garment geometry deformation process, and PhysGraph, which takes 0.057 seconds as it operates on the high-resolution garment meshes up-sampled from the coarse input.
All running times for MD and the detail enhancement methods were tested in the same computational environment: a 12th generation Intel Core i7 and a GeForce RTX 3090.

\begin{table}[h!]
    \caption{We report the computation time required to simulate a single frame of high-resolution garment geometry for Dress (C), (D), and (E), with 14,528, 27,855, and 43,257 vertices respectively. We compare the running efficiency of our GDSR method with high-resolution physics-based simulation in Marvelous Designer (MD), as well as with DDE \cite{zhang2021deep} and PhysGraph \cite{halimi2023physgraph}. The reported times are in seconds. 
    {For GDSR, PhysGraph and DDE, we report the algorithm running time plus ('+') the time cost of low-resolution simulation in MD.}}
     \begin{tabular}{c||c|c|c|c}
     \hline
        Dress & PBS & GDSR & PhysGraph & DDE \\
        \hline
        (C) & 0.194 &  \textbf{0.014}+0.045 & 0.057+0.045 &  0.202+0.045\\
        \hline
        (D) & 0.374 & \textbf{0.024}+0.058 & - & - \\
        \hline
        (E) & 0.559 & \textbf{0.032}+0.083 & - & - \\
        \hline
    \end{tabular}
    \label{tab:time-cost}
\end{table}

\clearpage
%
\appendix

\end{document}